\documentclass[runningheads]{llncs}
\usepackage[T1]{fontenc}
\usepackage{graphicx}
\usepackage{times}
\usepackage{soul}
\usepackage{url}
\usepackage[hidelinks]{hyperref}
\usepackage[utf8]{inputenc}
\usepackage[small]{caption}
\usepackage{graphicx}
\usepackage{amsmath}
\usepackage{booktabs}
\usepackage{algorithm}
\usepackage{algorithmic}
\usepackage[misc]{ifsym}
\usepackage{microtype}

\begin{document}
\title{An Ion Exchange Mechanism Inspired Story Ending Generator for Different Characters}
\titlerunning{Ion Exchange Mechanism for Story Ending Generation}
%
\author{Xinyu Jiang\textsuperscript{1}\thanks{The first two authors contribute equally to this work.}\and
Qi Zhang\inst{2,3}$^\star$ \and
Chongyang Shi~\Letter\inst{1} \and
Kaiying Jiang\inst{4} \and
Liang Hu\inst{3,5} \and
Shoujin Wang\inst{6}
}

\institute{Beijing Institute of Technology
\and University of Technology Sydney
\and Deepblue Academy of Sciences
\and University of Science and Technology Beijing
\and Tongji University
\and Macquarie University\\
\email{jiangxy\_08@163.com,\{zhangqi\_cs,cy\_shi\}@bit.edu.cn}}
\authorrunning{Xinyu Jiang, Qi Zhang, Chongyang Shi, Kaiying Jiang, Liang Hu, Shoujin Wang}
\maketitle              
\begin{abstract}
Story ending generation aims at generating reasonable endings for a given story context. Most existing studies in this area focus on generating coherent or diversified story endings, while they ignore that different characters may lead to different endings for a given story. In this paper, we propose a Character-oriented Story Ending Generator (CoSEG) to customize an ending for each character in a story. Specifically, we first propose a character modeling module to learn the personalities of characters from their descriptive experiences extracted from the story context. Then, inspired by the ion exchange mechanism in chemical reactions, we design a novel vector breaking/forming module to learn the intrinsic interactions between each character and the corresponding context through an analogical information exchange procedure. Finally, we leverage the attention mechanism to learn effective character-specific interactions and feed each interaction into a decoder to generate character-orient endings. Extensive experimental results and case studies demonstrate that CoSEG achieves significant improvements in the quality of generated endings compared with state-of-the-art methods, and it effectively customizes the endings for different characters.

\end{abstract}
\section{Introduction}
Story ending generation aims to deliver a comprehensive understanding of the context and predict the next plot for a given story \cite{DBLP:conf/aaai/GuanWH19,DBLP:conf/acl/LuoDYLCSS19,DBLP:conf/emnlp/XuRZZC018,DBLP:conf/ijcai/Wang019b}.
Some studies in this field generate coherent stories by modeling the sequence of events or verbs \cite{DBLP:conf/acl/FanLD19,DBLP:conf/aaai/MartinAWHSHR18}, or diversify story generation by introducing common senses or vocabulary information \cite{DBLP:conf/aaai/GaoBLLS19,DBLP:conf/emnlp/MaoMMC19}. Others focus on controlling the sentiment of story endings \cite{DBLP:conf/acl/LuoDYLCSS19,DBLP:conf/emnlp/TuDYG19} or generating the missing plot for an incomplete story \cite{DBLP:conf/aaai/ChenCY19,DBLP:conf/ijcai/Wang019b}. These methods generally ignore the relation and interaction between story plots and characters and simplify the influence of character personality on story generation, leading to desirable but character-irrelevant story endings.

\begin{figure}[t]
	\centering
	\includegraphics[width=0.8\columnwidth]{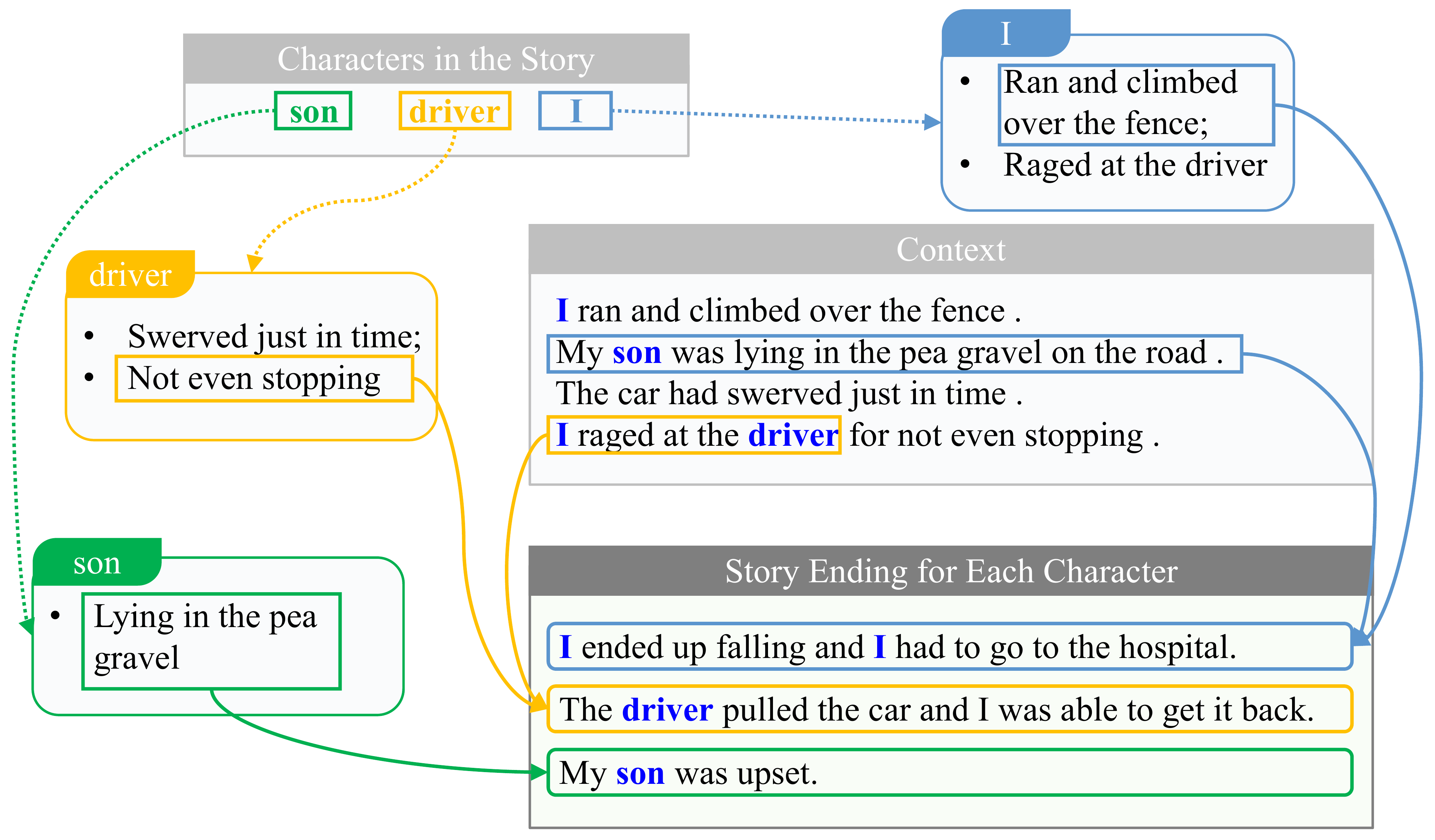}
	\caption{An example of the story context in the ROCStories corpus, and the endings generated by our model for different characters.}
	\label{Figure_0}
	\vspace{-2mm}
\end{figure}

Intuitively, stories are derived from characters, and character personality directly determines the plot and direction of the story. Figure~\ref{Figure_0} shows an example of a typical story in the ROCStories corpus \cite{DBLP:conf/repeval/MostafazadehVYK16} and the endings generated for different characters. From the figure, we can observe that: \textbf{1)} each character has its unique personality depicted by its character token and character experience, i.e., the character-related descriptions in a story. For example, the character \textbf{son} is depicted by the token “son” and the description “lying in the pea gravel”;
\textbf{2)} naturally, different characters with different personalities interact with the story context and thus affect the story plot, leading to different story endings (see the different endings for 'son', 'driver' and 'I' in the example).

Customizing the endings for different characters in a story is a novel but challenging task since there is no one-to-many dataset (i.e., one story corresponds to many ground-truth endings). To the best of our knowledge, most previous methods for the story ending generation aim to generate a single ending or missing plot rather than diverse coherent endings of different characters, for a given story context~\cite{DBLP:conf/aaai/GuanWH19,DBLP:conf/ijcai/Wang019b,DBLP:conf/emnlp/XuRZZC018}. The main challenges in customizing character-oriented story endings are 1) \emph{to model the personality of each character}, and 2) \emph{to learn the diverse interactions between different characters and the story context}.
Intuitively, a story context contains a character’s experiences, i.e., the multiple descriptions of the character, which depict the personalities of the character. It would be helpful to extract the related descriptions of each character from the story content and build its experience sequence via organizing the descriptions in chronological order for modeling each character's personalities.

\begin{figure*}[t]
	\centering
	\includegraphics[width=0.9\textwidth]{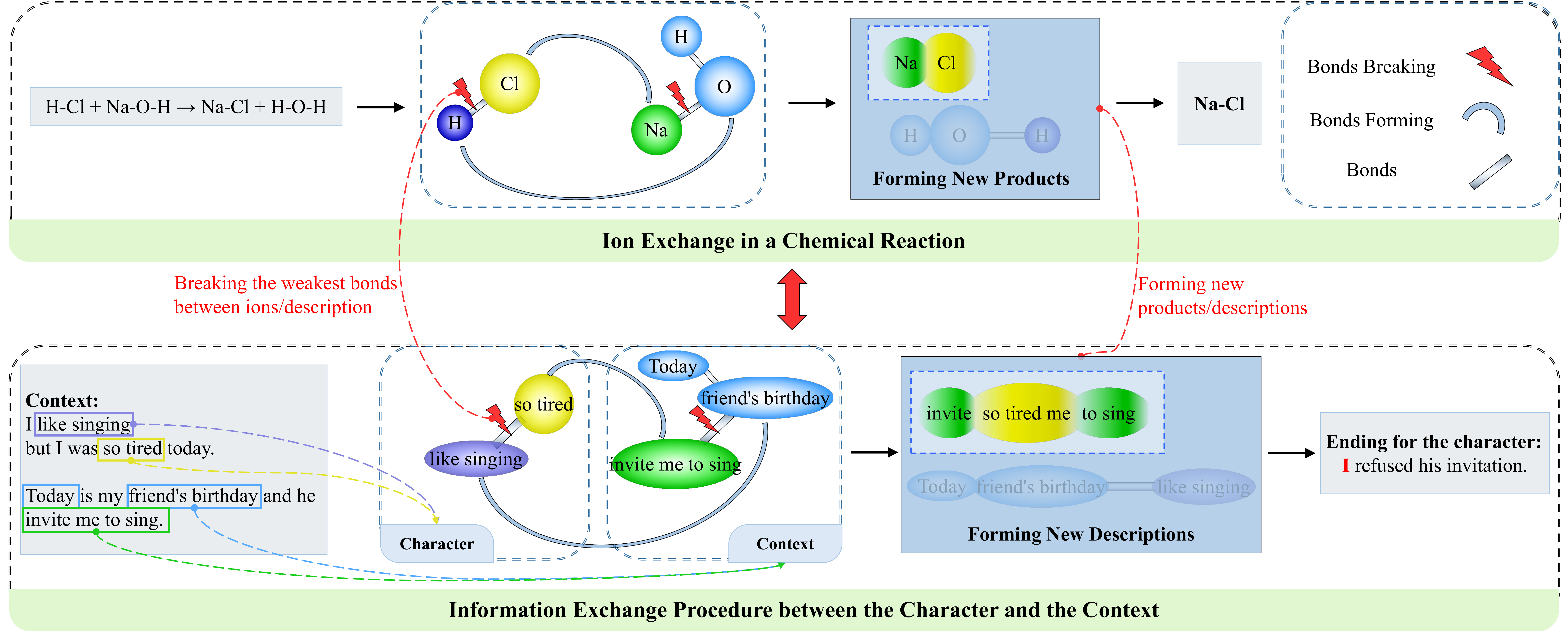}
	\caption{A character-context information exchange mechanism to learn the interaction between character and context, which is inspired by the ion exchange mechanism in chemical reactions.}
	\label{Figure_01}
\end{figure*}

Inspired by recent studies using deep learning to plan and predict chemical reactions~\cite{2018Planning,Sch2019Unifying}, we model the personalities of characters by analogizing the interactions between characters and context to chemical reactions. Specifically, we believe that the information exchange between different characters and context in generating new situational (character-specific) semantics during the interaction is similar to the ion exchange~\cite{2012Ion} to form new products in chemical reactions (cf. Figure~\ref{Figure_01}). Derived from this observation, it is promising to learn the interaction between a character and the corresponding context following an information exchange procedure. As depicted in Figure~\ref{Figure_01}, by exchanging related descriptions of the character and context, a new informative and character-related description “invite so tired me to sing” is formed by putting “so tired” (from character) and “invite me to sing” (from context) together. Consequently, the newly formed character-related description leads to an ending “refused his invitation” customized for the character “I” with a high probability.

Accordingly, we propose a Character-oriented Story Ending Generator (CoSEG) to customize an ending for each character in a story. Specifically, the proposed model first learns a representation of each character's personality by modeling its experiences with a Character Modeling module (CMM) and a context representation by modeling the story content. Then, a novel Vector Breaking/Forming module (VBF) is introduced to effectively learn the interaction between each character and the context through multiple information exchanges. Finally, a character-specific interaction representation is generated by adaptively picking out the most effective interaction via a Character-Context Attention module (C-CA), and each interaction representation is further utilized to customize the ending for the corresponding character. Note that CoSEG adopts LSTM-based encoder-decoder architecture, and the proposed key modules are network-agnostics and are also suitable for other prevailing networks, e.g., CNN~\cite{Kim14} and Transformer~\cite{XuPSPFAC20}. The main contributions of our paper are summarized below:
\begin{itemize}
	\item We propose a Character-oriented Story Ending Generator (CoSEG) model to tackle the challenging task of customizing story endings for different characters.
	\item We introduce a character modeling module to effectively model the personality of each character and learn a personalized and informative character representation.
	\item Inspired by the ion exchange process in chemical reactions, we propose a novel VBF module to learn the interaction between the character and context based on the information exchange mechanism.
\end{itemize}
Extensive experimental results on the ROCStories dataset show that our proposed CoSEG not only generates more coherent and diversified story endings compared with state-of-the-art and/or representative baseline methods, but also customizes effective endings for each character in a story. The superiority of CoSEG also demonstrates the effectiveness of the proposed CMM and VBF module in customizing story endings.

\section{Related Work}
Neural network-based models are the current mainstream in story generation methods owing to their impressive generation performance \cite{DBLP:conf/coling/MouSYL0J16,DBLP:conf/iclr/FedusGD18,DBLP:conf/emnlp/XuRZZC018,DBLP:conf/icml/WelleckBDC19,DBLP:conf/emnlp/MaoMMC19}. In recent years, there have been many innovations that utilize the encoder and decoder framework to generate coherent and diversified story endings.
\cite{DBLP:conf/naacl/YangYDHSH16} applies a hierarchical attention architecture to encode text information to generate the context representation. \cite{DBLP:conf/aaai/MartinAWHSHR18} predicts the next event by extracting the event represented from the sentence, thereby ensuring the coherence of the story.
\cite{DBLP:conf/acl/FanLD19} uses one head of the decoder's self-attention to attend only to previously generated verbs in order to generate a coherent story.
\cite{DBLP:conf/acl/LewisDF18} learns a second seq2seq model, which is led by the first model to focus on what the first model failed to learn. \cite{DBLP:conf/aaai/GuanWH19} introduces external knowledge and utilizes an incremental encoding scheme to ensure the diversity of stories. In addition, recent work proposes a character-centric neural storytelling model to generate stories for a given character~\cite{DBLP:conf/aaai/LiuLYHL0020}. Excited with the excellent performance of attention-based models~\cite{9404857,DBLP:journals/kbs/FengSHZJY21} like Transformer~\cite{VaswaniSPUJGKP17} and BERT~\cite{DevlinCLT19} in recent years, many story-ending generation and completion models leverage self-attention mechanism and Transformer architecture to enhance the quality of generated story endings~\cite{GuanHHZZ20,XuPSPFAC20}. In this work, we adopt LSTM as the backbone in the model design and experiments. Other architectures, e.g., attention networks and Transformer, can easily be incorporated into the proposed CoSEG. 

However, most of the aforementioned generation methods cannot generate multiple coherent and diverse endings for a single story context. Moreover, only a few works focus on generating multiple endings or responses given a single context.
\cite{DBLP:conf/aaai/GaoBLLS19} uses several unobserved latent variables $z$ to generate different responses. This method, however, relies on a one-to-many dataset. \cite{DBLP:conf/acl/LuoDYLCSS19} applies an additional sentiment analyzer to first predict the sentiment intensity $s$ of the ground truth ending $y$, then constructs paired data $(x,s;y)$ for training, where $x$ is the story context. In the generation stage, the model receives the sentiment variable $s$ from users to generate a sentiment-specific ending.
Recent work \cite{DBLP:conf/ijcai/Wang019b,DBLP:journals/ijon/MoWHCLZL21} introduce prevalent Transformers to learn story representation for generating a missing plot or a story ending. All these methods assume that the plot has little relation or interaction with the personality of the characters in the story.
Unlike these models, our proposed model can customize an ending for each character in the story context without relying on a one-to-many dataset.

\section{Character-oriented Story Ending Generator (CoSEG)}
\subsection{Problem Definition and Architecture}
In this section, we formulate the task of customizing the ending for each character in a story.
Given a story content $x=(x_1,...,x_l)$, which contains $l$ sentences, and $m$ characters $(c_1,...,c_m)$. The task is to predict customized endings $y=(y_1,...,y_m)$ for all the $m$ characters.

A story generally corresponds to only one ground truth ending, which may consist of the actions or opinions of a particular character; the endings of other characters are unavailable. To tackle the issue, in the training stage, we extract the experience sequence of the character in the ground-truth ending, then train our proposed model to generate an ending related to the extracted character (ground-truth ending). In the generating stage, we extract the experience sequences of all characters who appear in the story and then apply the proposed model to generate an ending for the characters.

\begin{figure*}[!t]
	\centering
	\includegraphics[width=1\textwidth]{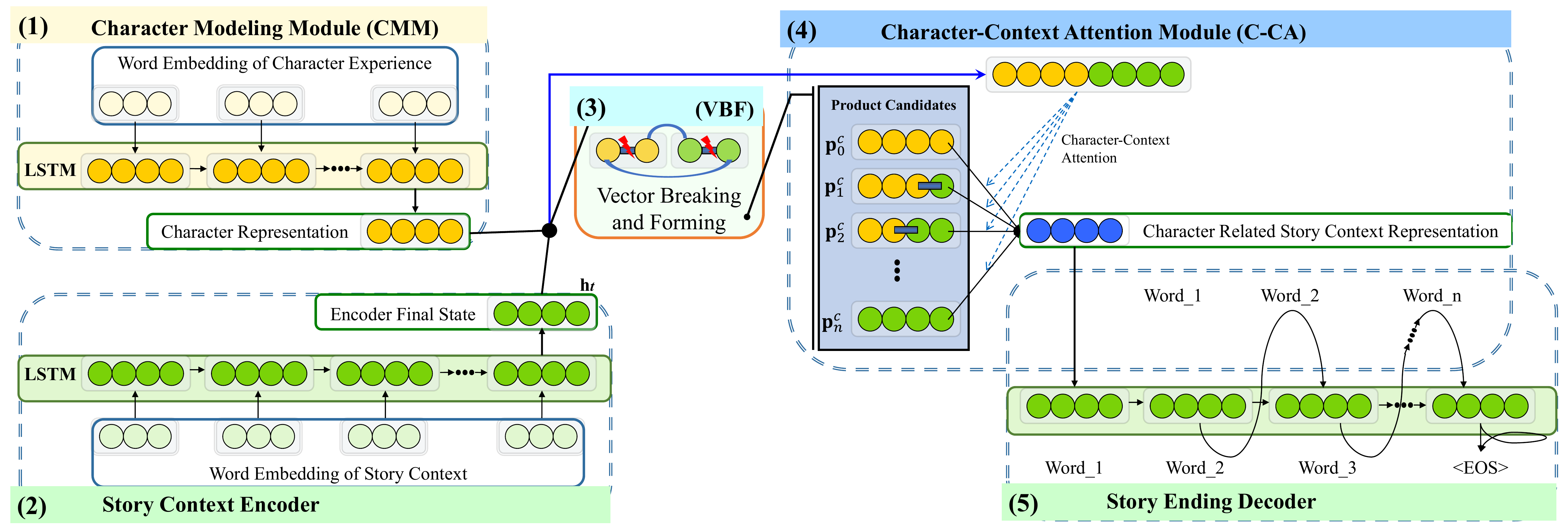}
	\caption{Character-oriented Story Ending Generator (CoSEG). }
	\label{Figure_1}
	\vspace{-2mm}
\end{figure*}

The architecture of our proposed CoSEG model is depicted in Figure~\ref{Figure_1}. Our model consists of three modules—a CMM module, a VBF module and a C-CA module—as well as an encoder to encode the story context and a decoder to generate the story ending. As shown in Figure~\ref{Figure_1}, the CMM module generates the character representation $cc$ for each character $c_i$ by modeling the character's experiences; the VBF module learns the interaction between the character representation and the story context through multiple information exchanges and generates multiple interaction results, namely product candidates $(\textbf{p}^c_0,...,\textbf{p}^c_n)$; the C-CA module generates a character-related story context representation by picking out the most effective product candidates. The context representation is further used as the initial state of the decoder to predict an ending for the character. The following sections present the details of each module.

\subsection{Character Experience Sequences}
We construct the experience sequence for each character in the story; the experience sequence is further fed into CMM to generate the character representation.

We take the sentence \emph{She knew a discount store near her sold socks} as an example to illustrate how to extract a character experience with the following four steps:

1) Construct a \textbf{dependency tree} for the given sentence, and get the headword \emph{knew} of the character \emph{She}.

2) Extract the \textbf{context words}, namely \emph{knew, discount, store, sold, socks}, as the first part of the character experience. This part is the background of the story.

3) Extract the \textbf{entity words}, namely the character \emph{She}, the headword \emph{knew} and the corresponding object \emph{store}, as the second part of the character experience. 

4) Connect the above two parts of character experience with a token \emph{OBJ} to obtain the final character experience $[$ \emph{knew, discount, store, sold, socks, OBJ, knew, store, She} $]$. The token \emph{OBJ} is utilized to separate the two parts, explicitly telling the model which words are context information and which are directly related to the character.

In this way, we can extract each character's experience from each sentence. Subsequently, we build the experience sequence $(e_1^i,...,e_{s}^i)$ of the character $c_i$ via organizing the character's experience in chronological order, where $s$ is the number of the sentences that contain $c_i$.

\subsection{Character Modeling Module (CMM)}
We generate the character representation by modeling the character experience sequence. Formally, given a character $c_i$ and a corresponding character experience sequence $(e_1^i,...,e_{S}^i)$, the module encodes the token sequence $(w_1^s,...,w_{T_s}^s)$ inside each experience $e_s^i$ to obtain the hidden states $(\textbf{h}_1^s,...,\textbf{h}_{T_s}^s)$, where $T_s$ is the length of $e_s^i$, and the superscript $s$ of $\textbf{h}$ or $w$ represents it is for the $s^{th}$ character experience.

We choose the final hidden state $\textbf{h}_{T_s}^s$ as the representation of the character $c_i$ who went through the character experience $e_s^i$.
The character representation $\textbf{h}_{T_s}^s$ is then used as the initial state of the next encoder to generate further enriched character representation $\textbf{h}_{T_{s+1}}^{s+1}$ as follows:
\begin{equation}
\textbf{h}_t^{s+1} = \textbf{LSTM}(\textbf{h}_{t-1}^{s+1}, \textbf{w}_t^{s+1}, \textbf{h}_{T_s}^s),
\end{equation}
where $T_{s+1}$ is the length of experience $e_{s+1}^i$, $w_t^{s+1}$ is the $t^{th}$ token in the sequence $(w_1^{s+1},...,w_{T_{s+1}}^{s+1})$ inside the experience $e_{s+1}^i$, $\textbf{w}_t^{s+1}$ denotes the embedding of $w_t^{s+1}$ and the superscript $s+1$ of $\textbf{h}$ or $w$ represents it is for the $(s+1)^{th}$ character experience.

Finally, the CMM Module will generate the $S^{th}$ character representation $\textbf{h}_{T_{S}}$, which has gone through all the character experiences $(e_1^i,...,e_{S}^i)$.

\subsection{Vector Breaking and Forming (VBF)}
Since both character and context are represented with high-dimensional vectors, we propose a novel VBF module to learn the interaction between the character and the story context representation based on the information exchange procedure (cf. Figure~\ref{Figure_01}) and then generate multiple product candidates.

\textbf{Vector Breaking:} Assume that there are invisible bonds between the adjacent elements of a vector, and VBF breaks the bonds between adjacent elements. For a vector of size $n$, there are $n+1$ potential bond-breaking positions. For example, given a vector $v_1=[0.1,0.2]$, the size of $v_1$ is 2 and we have 3 bond-breaking positions, as follows:
\begin{equation}
\begin{aligned}
v_1^l, v_1^r=[],[0.1,0.2]=VecB_{0}(v_1),\\
v_1^{l'}, v_1^{r'}=[0.1],[0.2]=VecB_{1}(v_1),\\
v_1^{l''}, v_1^{r''}=[0.1,0.2],[]=VecB_{2}(v_1),
\end{aligned}
\end{equation}
where $VecB_{k}$ represents breaking the bond in the position $k$, and the superscripts $l$ and $r$ of the breaking results represent the \emph{left} and \emph{right} parts of $v_1$, respectively.

\textbf{Vector Forming:} The two interaction vectors break at each position respectively. To keep the size of interaction results constant, for each bond-breaking position, VBF integrates the left part of the first vector and the right part of the second vector to generate a product candidate\footnote{There is no order between the two interaction vectors, which vector as the first one has little influence on the experimental results.}. In this way, two vectors of size $n$ can interact to obtain a total of $n+1$ product candidates. For example, let $v_1$ interact with $v_2=[0.3,0.4]$, and we can obtain such three product candidates $\textbf{p}^c_0=[0.3,0.4], \textbf{p}^c_1=[0.1,0.4], \textbf{p}^c_2=[0.1,0.2]$. As shown in Figure~\ref{Figure_1} part (3), the character representation and the encoder final state $\textbf{h}_t$ (i.e., story context representation) interact in the VBF module to generate the product candidates $(\textbf{p}^c_0,...,\textbf{p}^c_n)$.

\subsection{Character-Context Attention (C-CA)}
As shown in Figure~\ref{Figure_1} part (4), the C-CA Module aims to pick out the most effective product candidates. Specifically, we utilize the $s^{th}$ character representation $\textbf{h}_{T_s}$ and the encoder final state $\textbf{h}_t$ to obtain the attention weight of each product candidate $\textbf{p}^c_k$:
\begin{equation}
\begin{aligned}
\textbf{a}^s&=\sigma (\textbf{W}_{a}[\textbf{h}_{T_s};\textbf{h}_t] + \textbf{b}_{a}),\\
\textbf{r}^s&=\sum_{k=0}^{n}\textbf{a}_k^s\textbf{p}^c_k,
\end{aligned}
\end{equation}
where $\sigma$ is the softmax function, $\textbf{W}_{a}$ is the weight matrix, $\textbf{b}_{a}$ is the bias, $att$ is the attention weight, and $\textbf{a}^s$ stands for the character-related story context representation of the $s^{th}$ character. As shown in Figure~\ref{Figure_1} part (5), the $\textbf{r}^s$ is further used as the initial state of the decoder (note that we omit the superscript $s$ in the following for simplicity):
\begin{equation}
\textbf{h}_t^y = \textbf{LSTM}(\textbf{h}_{t-1}^y, \textbf{w}_{t-1}^y, \textbf{r}),
\end{equation}
where $\textbf{h}_t^y$ is the $t^{th}$ hidden state of the decoder, which is further utilized to generate the $t^{th}$ word $w_{t}^y$, $\textbf{w}_{t-1}^y$ is the embedding of word $w_{t-1}^y$.

\section{Experiment}
In this section, we conduct extensive experiments to investigate the quality of CoSEG and the comparative baselines.
\subsection{Dataset}
We evaluated our model on the ROCStories corpus \cite{DBLP:conf/repeval/MostafazadehVYK16}. This corpus contains 98,162 five-sentence stories. Our task is to generate an ending for each character that appears in a given four-sentence story\footnote{We identify characters in a macro way. We extract the subject of each sentence in the story. We regard \textit{name entity} or \textit{noun} as the character of the sentence. In principle, in this way, we can generate an ending for any \textit{noun}. Since there are few endings regarding nouns as the characters in the training data (such as the ending with "car" as the character), the proposed model is difficult to generate high-quality endings for those characters.}.
For each story, we extract the experience sequence for the character who appears in the ground truth ending. We select 66,881 stories in which the length of the ground-truth character's experience sequence is no less than 2 and treat these stories as the training set. We elaborately design two test sets, each with 3073 stories. Specifically, the two sets are called sufficient test set and inadequate test set. In the sufficient test set, the length of the ground-truth character's experience sequence is no less than 2 for all stories, while the length is less than 2 for all stories in the inadequate test set. The two test sets are applied to evaluate the performance of our proposed model when the character information is sufficient and inadequate, respectively.

\subsection{Experimental Settings}
We use the GloVe.6B \cite{DBLP:conf/emnlp/PenningtonSM14} pre-trained word embedding, and the number of dimensions is 200. The hidden size of the LSTM cell is 512. Since the size of the character representation and the encoder's final state both are 512, the number of product candidates will be $512+1=513$. A larger dimension size brings large computation costs for the model and the device. In summary, the detailed experimental settings are provided as follows:

\begin{itemize}
	\item We use 66,881 stories for training.
	\item We use 3073 stories for validation, which is shared by the two test sets.
	\item We have two test sets with 3073 stories each.
	We refer to the two test sets as the sufficient test set and the inadequate test set respectively.
	In the sufficient test set, the character in the ground truth ending also appears multiple times in the story context, and in the inadequate test set is the opposite.
	These two test sets evaluate the performance of our proposed model when the character information is sufficient and inadequate, respectively.
	\item We use Momentum Optimizer to update parameters when training and empirically set the momentum to be 0.9.
	\item The size of the character representation and the encoder hidden state are 512.
	\item We select the product candidates generated using $VecB_{0}$, $VecB_{128}$, $VecB_{256}$, $VecB_{384}$ and $VecB_{512}$ five breaking operation.	
	\item The number of product candidates will be $512+1=513$. 
\end{itemize}
Note that the reason for the selection of the product candidates is analyzed in \textit{Combination Analysis on Product Candidates} in Section \ref{param}.

\subsection{Baselines}
We compared our model with the following state-of-the-art baseline methods:

\textbf{Seq2Seq} \cite{DBLP:conf/emnlp/LuongPM15}: A vanilla encoder-decoder model with an attention mechanism. The model treats the story context as a single sentence.

\textbf{HAN} \cite{DBLP:conf/naacl/YangYDHSH16}: A hierarchical attention architecture is applied to encode text information so as to generate the context representation.

\textbf{IE} \cite{DBLP:conf/aaai/GuanWH19}: It adopts an incremental encoding scheme to represent context clues and applies commonsense knowledge by multi-source attention.

\textbf{T-CVAE} \cite{DBLP:conf/ijcai/Wang019b}: It proposes a conditional variational autoencoder based on Transformer for missing plot generation.

\textbf{MGCN-DP} \cite{DBLP:conf/aaai/HuangMLCLWLL21}:
It leverages multi-level graph convolutional networks over dependency parse trees to capture dependency relations and context clues.

In addition, we introduce two variations of the proposed CoSEG model:

\textbf{CoADD}: We replace the VBF Module in CoSEG with an element-wise summation.

\textbf{CoCAT}: We concatenate the character representation and the encoder final state, and pass the concatenated vector through a linear layer to obtain the character-related story context representation.

\subsection{Evaluation Metrics}
We evaluate our model from two perspectives: the quality of the generated endings and the ability to customize endings.

\subsubsection{Quality Evaluation} We adopt two kinds of evaluations to investigate the ability of the proposed method and the baselines in generating high-quality story endings.

\textbf{Automatic Evaluation:} We use perplexity (PPL) and BLEU (BLEU-1, BLEU-2 and BLEU-3) \cite{DBLP:conf/acl/PapineniRWZ02} to evaluate the quality of the generated endings. A smaller PPL and a higher BLEU indicate a better ending.

\textbf{Manual Evaluation:} We hire three evaluators, who are experts in English, to evaluate the generated story endings.
We randomly sampled 200 stories from the two test sets and obtained 1400 endings from the seven models for each test set.
Evaluators need to score the generated endings in terms of two criteria: coherency and grammar.
The coherency score measures whether the endings are coherent with the story context; specifically, the score of 3 denotes coherency, the score of 1 denotes coherency to some extent, and the score of 0 denotes no coherency at all. In addition, the grammar score measures whether there are grammatical errors in generated endings; a grammar score is 0 if endings have errors, and 1 otherwise.

\subsubsection{Ability to Customize Endings}
We randomly sample 200 stories from the two test sets and generate ending for one random character in each story. To evaluate the ability to customize endings for different characters, we propose three evaluation metrics:

\textbf{Success Rate (SucR):} SucR measures whether the subject of the generated ending is the selected character.

\textbf{Rationality:} We adopt three levels to evaluate whether the generated ending matches the selected character given the story context: level 3 denotes perfect matching, level 1 denotes partial matching, and level 0 for mismatching.

\textbf{Discrimination Degree (DiscD):}
Given an ending generated by our proposed model, we further hire three evaluators to choose which character is the ending generated for. If the character chosen by the evaluator is consistent with the selected character, it scores 1; and 0 otherwise.

\subsection{Evaluation Results}
\label{param}
\begin{table}[!t]
	\caption{Automatic evaluation results of the sufficient and the inadequate test set.}
	\label{Table_1}
	\centering
	\small
	\begin{tabular}{lcccc}
		\toprule
		\multicolumn{5}{c}{Sufficient}\\
		\midrule
		Model & PPL & BLEU-1 (\%) & BLEU-2 (\%) & BLEU-3 (\%) \\
		\midrule
		Seq2Seq    & 13.26& 22.46 & 6.88 & 4.21 \\
		HAN    & 13.43& 22.43 & 6.96 & 4.47 \\
		IE    & 12.08& 23.08 & 7.43 & 4.67 \\
		T-CVAE    & 11.21& 23.72 & 8.05 & 5.11 \\
		MGCN-DP    & 11.01 & 23.90 & 8.11 & 5.34 \\
		\midrule
		CoADD & 12.14 & 23.92 & 7.74 & 4.68 \\
		CoCAT & 11.45 & 24.26 & 8.53 & 5.41 \\
		CoSEG & \textbf{9.99} & \textbf{25.28} & \textbf{9.10} & \textbf{5.93}\\
		\midrule
		\multicolumn{5}{c}{Inadequate}\\
		\midrule
		Model & PPL & BLEU-1 (\%) & BLEU-2 (\%) & BLEU-3 (\%) \\
		\midrule
		Seq2Seq    & 21.81& 17.13 & 3.76 & 1.76 \\
		HAN    & 24.26& 17.15 & 4.08 & 2.32 \\
		IE    & 16.90& 18.40 & 4.89 & 2.78 \\
		T-CVAE    & 17.08& 22.10 & 7.05 & 4.22 \\
		MGCN-DP    & 18.16 & 20.89 & 5.90 & 3.68 \\
		\midrule
		CoADD & 14.53 & 21.83 & 6.99 & 4.03 \\
		CoCAT & 15.08 & 24.50 & 9.09 & 5.26 \\
		CoSEG & \textbf{11.45} & \textbf{26.06} & \textbf{9.80} & \textbf{5.70}\\
		\bottomrule
	\end{tabular}
\end{table}

\noindent\textbf{Automatic Evaluation} The automatic evaluation results for the sufficient and inadequate test sets are shown in Table~\ref{Table_1}.
From the table, we can observe the following:

1) In both the sufficient and inadequate test sets, our model has lower perplexity and higher BLEU scores than the baselines.
Specifically, in terms of perplexity, \textbf{CoSEG} outperforms \textbf{MGCN-DP}, \textbf{T-CVAE}, \textbf{IE}, \textbf{HAN} and \textbf{Seq2Seq} by 1.02/ 1.22/ 2.09/ 3.44/ 3.27 respectively in the sufficient test set, and by 6.01/ 5.63/ 5.45/ 12.81/ 10.36 respectively in the inadequate test set.
In addition, in terms of BLEU-1, \textbf{CoSEG} outperforms \textbf{MGCN-DP}, \textbf{T-CVAE}, \textbf{IE}, \textbf{HAN} and \textbf{Seq2Seq} by 1.38\%/ 1.56\%/ 2.2\%/ 2.85\%/ 2.82\% respectively in the sufficient test set, and by 5.17\%/ 3.96\%/ 7.66\%/ 8.91\%/ 8.93\% respectively in the inadequate test set.

2) Our \textbf{CoSEG} model has the smallest performance gap between the two test sets, which illustrates the performance of our model is not easily affected by the amount of information. Specifically, the perplexity increased by 1.55 in the inadequate test set based on the sufficient test set, and the BLEU-1 increased by 0.78\%.

3) In both the sufficient and inadequate test set, the \textbf{CoSEG} model outperforms the \textbf{CoADD} and the \textbf{CoCAT} a lot, which illustrates the interaction ability of the VBF Module is much stronger than the addition and concatenation.

\begin{table}[ht]
	\caption{Manual evaluation results of the sufficient and the inadequate test set.}
	\label{Table_2}
	\centering
	\small
	\begin{tabular}{lcccc}
		\toprule
		& \multicolumn{2}{c}{Sufficient} & \multicolumn{2}{c}{Inadequate}\\
		\midrule
		Model & Coherency & Grammar & Coherency & Grammar\\
		\midrule
		Seq2Seq    & 1.395& 0.655 & 0.905& 0.780 \\
		HAN    & 1.160& 0.685  & 0.600& 0.785\\
		IE    & 1.360& 0.760 & 1.210& 0.820\\
		T-CVAE    & 1.750& 0.785 & 1.440& 0.815\\
		MGCN-DP    & 1.760 & 0.780  & 1.315 &0.795 \\
		\midrule
		CoADD & 1.690 & 0.775 & 1.220 & 0.760\\
		CoCAT& 1.855 & 0.605 & 0.965 & 0.705\\
		CoSEG & \textbf{1.880} & \textbf{0.805} & \textbf{1.620} & \textbf{0.835}\\
		\bottomrule
	\end{tabular}
	\vspace{-1mm}
\end{table}

\textbf{Manual Evaluation} The manual evaluation results for the sufficient and inadequate test sets are shown in Table~\ref{Table_2}, where we can observe:

In both the sufficient and inadequate test set, the \textbf{CoSEG} model obtains the best coherency score and the best grammar score.
Specifically, in terms of Coherency, \textbf{CoSEG} outperforms \textbf{CoCAT}, \textbf{CoADD}, \textbf{MGCN-DP},\textbf{T-CVAE}, \textbf{IE}, \textbf{HAN} and \textbf{Seq2Seq} by 0.025/ 0.19/ 0.12/ 0.13/ 0.52/ 0.72/ 0.485 respectively in sufficient test set, and by 0.655/ 0.4/ 0.305/ 0.18/ 0.41/ 1.02/ 0.715 respectively in inadequate test set.
Moreover, in terms of Grammar, \textbf{CoSEG} outperforms \textbf{CoCAT}, \textbf{CoADD}, \textbf{MGCN-DP}, \textbf{T-CVAE}, \textbf{IE}, \textbf{HAN} and \textbf{Seq2Seq} by 0.2/ 0.03/ 0.025/ 0.02/ 0.045/ 0.12/ 0.15 respectively in sufficient test set, and by 0.13/ 0.075/ 0.04/ 0.02/ 0.015/ 0.05/ 0.055 respectively in inadequate test set.
\begin{table}[ht]
    \vspace{-6mm}
	\caption{Ability to Customize Endings.}
	\label{Table5}
	\centering
	\small
	\begin{tabular}{lccc}
		\toprule
		Testset & SucR & Rationality & DiscD \\
		\midrule
		Sufficient & 0.855 & 1.965 & 0.755\\
		Inadequate & 0.605 & 1.980 & 0.555\\
		\bottomrule
	\end{tabular}
	\vspace{-2mm}
\end{table}

\noindent\textbf{Ability to Customize Endings} 
The ability to customize endings for different characters of our model is shown in Table~\ref{Table5}, where we can observe:

1) In the sufficient test set, the success rate (SucR) and the discrimination degree (DiscD) are 85.5\% and 75.5\% respectively, which indicates that our model is able to identify the differences between characters. The SucR and DiscD in the inadequate test set are lower than there in the sufficient test set, which illustrates that the amount of character information has a certain influence on distinguishing different characters.

2) In both the sufficient and inadequate test sets, the rationality of the customized endings is close to 2.0 on the premise that the maximum score is 3.0. It indicates that our model has a high probability of 66\% to predict a reasonable ending for each character.

\begin{table}[!t]
	\caption{Experimental results of several different combinations.}
	\label{Table_A1}
	\begin{center}
		\begin{small}
			\begin{tabular}{lc}
				\toprule
				Model & PPL \\
				\midrule
				CoSEG (0)   & 12.08 \\
				CoSEG (128)   &  14.74\\
				CoSEG (256)   & 13.06 \\
				CoSEG (0-256-512)  & 10.93\\
				CoSEG (0-128-256-384-512) & \textbf{9.99}\\
				\bottomrule
			\end{tabular}
		\end{small}
	\end{center}
	\vspace{-2mm}
\end{table}

\noindent\textbf{Combination Analysis on Product Candidates} We conduct experiments on several different combinations of product candidates. Specifically, the \textbf{CoSEG (n)} selects the product candidate generated using the $VecB_{n}$ breaking operation; the \textbf{CoSEG (0-256-512)} selects the product candidates generated using $VecB_{0}$, $VecB_{256}$ and $VecB_{512}$ three breaking operations; the \textbf{CoSEG (0-128-256-384-512)} selects the product candidates generated using $VecB_{0}$, $VecB_{128}$, $VecB_{256}$, $VecB_{384}$ and $VecB_{512}$ five breaking operations.
	
The experimental results are shown in Table~\ref{Table_A1}. We can observe that \textbf{CoSEG (0-128-256-384-512)} achieves the best performance. The result explains that we finally selected the product candidates generated using $VecB_{0}$, $VecB_{128}$, $VecB_{256}$, $VecB_{384}$ and $VecB_{512}$ five breaking operations and utilize the selected five product candidates as inputs to C-CA module. In addition, the result is attributed to the fact that \textbf{CoSEG (0-128-256-384-512)} involves more and smaller candidate which facilitate generating fine-grained semantic elements and providing more semantic combinations.

\begin{table}[!t]
	\caption{Case Study: Endings Generated by Different Models.}
	\label{Table_A6}
	\begin{center}
		\begin{small}
			\begin{tabular}{ll}
				\toprule
        		Context: &Ned was walking in the park one day. He noticed\\& the sky started to turn gray. Ned turned back toward\\& his house. He didn't quite get home before it started\\& raining. \\
        		Gold: & Ned ran inside, a little wet, but happy to be home.\\
        		\midrule
        		Seq2Seq:& He went back to his car and bought with his warm.\\
        		HAN:&He had a clean mess.\\
        		IE:&He had to go home and go home.\\
        		T-CVAE:&Ned looked around and saw the sky.\\
        		MGCN-DP:&he was very happy with his new car!\\
        		\midrule
        		CoADD: & He decided to go to the store to buy more umbrella.\\
        		CoCAT: & He went outside to find his car. \\
        		CoSEG: & He decided to go to the store to get an umbrella. \\
        		\midrule
        		\midrule
				Context: &We removed a lamp last week during carpet cleaning.\\
				&After the cleaners left \textbf{my wife} put the lamp back.\\
				&The lighting seemed much dimmer.\\
				&Then \textbf{she} realized she had brought out the wrong lamp.\\
				Ground truth: & I said I had not even noticed.\\
				\midrule
				Seq2Seq:& She was so relieved.\\
				HAN:&She had to clean it up.\\
				IE:&She had to clean it up and get a new one.\\
				T-CVAE:&I bought a new lamp .\\
				MGCN-DP:&We had to clean it up.\\
				\midrule
				CoADD: & I was so mad that she was able to get it fixed.\\
				CoCAT: & She was so happy to have a new lamp.\\
				\textbf{CoSEG}: & I was so \textbf{upset} , I \textbf{decided} to \textbf{get her} a new one.\\
				\midrule
				\midrule
				Context: &Larry \textbf{needed a job}.\\
				&He was let go from a job a month ago and hadn't found another one.\\
				&His \textbf{bills were starting to mount up}.\\
				&Larry was determined and applied to many jobs every day.\\
				Ground truth: & He was finally hired.\\
				\midrule
				Seq2Seq:& He was able to get a job in time and help for his job.\\
				HAN:&He was able to get a job in a new city.\\
				IE:&He was hired for a new job.\\
				T-CVAE:&He have a job.\\
				MGCN-DP:&He said he was going to be more careful.\\
				\midrule
				CoADD: & He was able to get to work and get a job.\\
				CoCAT: & He was hired at the store and bought a brand new car.\\
				\textbf{CoSEG}: & He finally \textbf{got a job} offer and \textbf{paid his} \textbf{bills}.\\
				\bottomrule
			\end{tabular}
		\end{small}
	\end{center}
\end{table}

\section{Case Study}
\subsection{Ground-Truth Endings}
We present several examples of ground-truth story endings generated by baselines and our model in Table~\ref{Table_A6} to demonstrate that our model is able to generate more natural and more character-related endings than the baselines. Specifically, in the first story in Table~\ref{Table_A6}, the ending generated by baseline \textbf{T-CVAE} makes the reader feel that the character is not smart enough, it has begun to \emph{rain}, and it should not be very useful to \emph{look at the sky} at this time; the baseline \textbf{IE} only describes \emph{He had to go home}, without other details. And other baselines predicted wrong endings. By contrast, our model \textbf{CoSEG} allows the character to take an appropriate action, \emph{get an umbrella}, according to the given story context, \emph{started raining}.

In addition, the second story in Table~\ref{Table_A6} is derived from the inadequate test set. In this example, the context is a story about a wrong lamp. The baselines \textbf{HAN}, \textbf{IE} and \textbf{MGCN-DP} describe \emph{She/We had to clean it up}, and \textbf{IE} further describes \emph{get a new one}. Our model \textbf{CoSEG} not only allows the character to have an appropriate emotion, \emph{I was so upset}, but also let the character take a reasonable action, \emph{decided to get her a new one}. Obviously, the ending generated by \textbf{CoSEG} takes the character's emotions (\emph{\textbf{upset}}), behaviors (\emph{\textbf{decided}...\textbf{get}...}), and relationships (\emph{get \textbf{her} a...}) into account, which illustrates the ability of our model to obtain character's personality.

The third example in Table~\ref{Table_A6} is sampled from the sufficient test set. In this example, the context is a story about a man's bills mount up and he needs a job. Most of the baselines describe \emph{He get a job}, as well as the ground truth and our proposed model. In addition, different from all baselines and the ground truth endings, our model further describes the man's purpose of looking for a job, \emph{\textbf{paid bills}}, which also demonstrates that our model is able to generate a more character-related ending.

\begin{table}[!t]
	\caption{Customizing endings for each character using our proposed CoSEG model.}
	\label{Table_A7}
	\begin{center}
		\begin{small}
			\begin{tabular}{ll}
				\toprule
				Context: & \textbf{I} ran and climbed over the fence.\\
				&\textbf{My son} was lying in the pea gravel on the road.\\
				&The car had swerved just in time.\\
				&\textbf{I} raged at the \textbf{driver} for not even stopping.\\
				Ground truth: & I called 911 to come get my child.\\
				\midrule
				\multicolumn{2}{l}{Endings for each character:}\\
				\midrule
				For \textbf{I}: & I ended up \textbf{falling} and I had to \textbf{go to the} \textbf{hospital}.\\
				For \textbf{driver}: & The driver \textbf{pulled the car} and I was able to get it back.\\
				For \textbf{son}: & My son was \textbf{upset}.\\
				\midrule
				\midrule
				Context: & \textbf{I} had a dental appointment I had to go to today.\\
				&While getting my teeth checked , my \textbf{dentist} told \textbf{I} had a cavity.\\
				&\textbf{He} said it's probably because \textbf{I}'ve been using subpar toothpaste.\\
				&\textbf{I}'ve been using the same toothpaste \textbf{he} recommended six months ago.\\
				Ground truth: & Thanks a lot for the recommendation, doc.\\
				\midrule
				\multicolumn{2}{l}{Endings for each character:}\\
				\midrule
				For \textbf{I}: & I am \textbf{glad} I \textbf{have a new toothpaste}.\\
				For \textbf{dentist}: & The dentist \textbf{told me} that he \textbf{had to get a} \textbf{new toothpaste}.\\
				\midrule
				\midrule
				Context: & \textbf{John} was awakened by a phone call.\\
				&Answering , \textbf{John} realized it was his buddy, \textbf{Rich}.\\
				&\textbf{Rich} said he was stranded on a highway just outside of town.\\
				&\textbf{John} drove out to pick up \textbf{Rich}.\\
				Ground truth: & John drove Rich home , where they both fell asleep on the couch.\\
				\midrule
				\multicolumn{2}{l}{Endings for each character:}\\
				\midrule
				For \textbf{Rich}: & He \textbf{drove to the mall} and \textbf{bought a new}  \textbf{car}. \\
				For \textbf{John}: & John and his friends \textbf{went to the park} and had a great time. \\
				\bottomrule
			\end{tabular}
		\end{small}
	\end{center}
\end{table}

\subsection{Character-Orient Endings}
In this section, we present three examples with character-orient endings (including the ground-truth endings) generated by our method in Table~\ref{Table_A7}, to illustrate the ability of our model to customize endings for different characters.

In the first example in Table~\ref{Table_A7}, our model customizes endings for the characters \emph{I}, \emph{driver} and \emph{son}. The context is a story about a car accident that happened to a father and son. The endings generated by our model are that the \emph{father} had to go to the hospital, the \emph{driver} ran away, and the \emph{son} was upset. These three endings generated by our model describe the behavior of the \emph{father} and the \emph{driver} and the mood of the \emph{son}, which demonstrate the effectiveness of our proposed model to customize endings for different characters. The second example in Table~\ref{Table_A7} is a story in that a man has a cavity because he uses the toothpaste recommended by the dentist, and our model customizes endings for the characters \emph{I} and \emph{dentist}. The third example in Table~\ref{Table_A7} is a story in that Rich is stranded on a highway and he calls John to pick him up. Our model identifies the differences between Rich and John, generating endings, \emph{He drove to the mall and bought a new car}, for Rich, and \emph{John and his friends went to the park and had a great time}, for John. Rich \emph{\textbf{need a new car}}, because his car is \emph{\textbf{stranded on the highway}}.

\section{Conclusion}
To tackle the challenging task of customizing story endings for different characters, we propose a Character-oriented Story Ending Generator (CoSEG). Experimental results demonstrate that our proposed model can not only generate more coherent and diversified story endings compared with state-of-the-art methods but also effectively customize the ending for each character in a story.

\section{Acknowledgements}
This work is supported by the National Key Research and Development Program of China(No. 2019YFB1406300), National Natural Science Foundation of China (No. 61502033) and the Fundamental Research Funds for the Central Universities.
%
%
%
 \bibliographystyle{splncs04}
 \bibliography{main}
 \nocite{*}
%


\end{document}